\def\year{2021}\relax
\documentclass[letterpaper]{article} 
\usepackage{aaai21}  
\usepackage{times}  
\usepackage{helvet} 
\usepackage{courier}  
\usepackage[hyphens]{url}  
\usepackage{graphicx} 
\usepackage{natbib}  
\usepackage{caption} 
\usepackage{amsmath}
\usepackage{amsfonts}
\usepackage{amssymb}
\usepackage{mathtools}
\usepackage{dsfont}
\usepackage{multirow}
\usepackage[switch,mathlines,displaymath]{lineno}
\DeclareMathOperator*{\argmax}{argmax}

\title{Unsupervised Domain Adaptation for Semantic Segmentation by Content Transfer}
\author{
    Suhyeon Lee,
    Junhyuk Hyun,
    Hongje Seong,
    Euntai Kim\thanks{Corresponding author}
    \\
}
\affiliations{
    School of Electrical and Electronic Engineering, Yonsei University, Seoul, Korea\\

    \{hyeon93, jhhyun, hjseong, etkim\}@yonsei.ac.kr

}

\begin{document}

\maketitle

\begin{abstract}
In this paper, we tackle the unsupervised domain adaptation (UDA) for semantic segmentation, which aims to segment the unlabeled real data using labeled synthetic data. The main problem of UDA for semantic segmentation relies on reducing the domain gap between the real image and synthetic image. To solve this problem, we focused on separating information in an image into content and style. Here, only the content has cues for semantic segmentation, and the style makes the domain gap. Thus, precise separation of content and style in an image leads to effect as supervision of real data even when learning with synthetic data. To make the best of this effect, we propose a zero-style loss. Even though we perfectly extract content for semantic segmentation in the real domain, another main challenge, the class imbalance problem, still exists in UDA for semantic segmentation. We address this problem by transferring the contents of tail classes from synthetic to real domain. Experimental results show that the proposed method achieves the state-of-the-art performance in semantic segmentation on the major two UDA settings.
\end{abstract}

\section{Introduction}
Semantic segmentation is a task of classifying each pixel of an image into semantic categories (e.g., sky, road, traffic sign, etc.). Training the segmentation model~\cite{chen2017deeplab} requires large amounts of pixel-level annotated data, but pixelwise image labeling is extremely labor-intensive and time-consuming. To reduce the labeling cost, the model learned with automatically labeled synthetic datasets~\cite{richter2016playing, ros2016synthia} can be used in real-world environments~\cite{cordts2016cityscapes}. In this case, the domain gap between synthetic data and real data causes the model to behave poorly in real-world environments, so unsupervised domain adaptation (UDA) methods have been proposed to solve this problem~\cite{chang2019all, du2019ssf, lin2017refinenet, luo2019taking}. 

UDA for semantic segmentation aims to train a segmentation model that performs on the real domain by using unlabeled real images and labeled synthetic images. Since the supervision of semantic information comes only from synthetic data, reducing the domain gap between the real image and synthetic image is a major problem. That is, if we completely adapt the real and synthetic domains, learning with synthetic data would be the same as learning with real data. We address this problem by forcing networks to extract content and style from an image separately.

An image includes content and style~\cite{gatys2016image, isola2017image}. The content has cues for semantic segmentation regardless of style. The style is a cue to represent a characteristic for a domain of an image regardless of content. The style makes a domain gap, so separating style from content further enhances the alignment between the two domains in the content space. However, separating content and style in an image is challenging because we cannot directly supervise what content or style is. In this paper, we introduce zero-style loss to capture content and style separately. To do this, we set one content encoder and two style encoders. By learning with our proposed zero-style loss, the content encoder and style encoders can capture domain-invariant content feature and domain-specific style features, respectively.

With the UDA for semantic segmentation, we can provide a large number of additional labeled synthetic images for training. However, this method overlooks a serious problem. In the urban scene dataset, the number of samples per class is not equally distributed. Head classes (e.g., road, sky) exist in most of the data, but tail classes (e.g., trafﬁc sign, bicycle) are rarely shown. The model learned with the class-imbalanced dataset would be biased toward head classes~\cite{cui2019class}.

To further boost the performance on the target (real) domain, the self-training strategy has recently been applied to the UDA methods~\cite{li2019bidirectional, french2017self}. This strategy generates the target pseudo labels using the segmentation model, which is trained with labeled source (synthetic) data. The class imbalance problem is further accelerated when using this self-training technique. Since the segmentation model is already biased toward the head classes, we cannot expect the correct generation of pseudo labels for the tail classes. Thus, self-training makes networks to be further biased toward head classes.

To solve this problem, we propose the tail-class content transfer. Since we trained our model to separate the content and style from an image, our model can extract the content from the source data, and this content would be usable in the target domain. Therefore, we transfer the tail-class contents from the source to the target. Note that unlike the target pseudo label generated by the segmentation model, our content transfer enables the model to supervise using accurately annotated source labels in the target domain.

To sum up, our contributions are as follows:
\begin{itemize}
\item We propose the zero-style loss that enables more precise feature-level domain adaptation and domain transfer by separating the content and style of the image independently.
\item We propose the content transfer to solve the class imbalance problem in UDA for semantic segmentation. This method is end-to-end trainable without pre-processing datasets.
\item We achieve the state-of-the-art performance for two synthetic-to-real UDA settings; GTA5 $\rightarrow$ Cityscapes and SYNTHIA $\rightarrow$ Cityscapes.
\end{itemize}

\section{Related Work}

\subsection{Semantic Segmentation}
Semantic segmentation is a computer vision task that predicts category per pixel of image. Recently, convolutional neural network-based methods have been developed. \citet{long2015fully} proposed Fully Convolutional Network (FCN) for spatially dense prediction. DeepLab~\cite{chen2017deeplab} proposed Atrous Spatial Pyramid Pooling (ASPP) to reliably segment on various scales. PSPNet~\cite{zhao2017pyramid} proposed pyramid pooling module (PPM) that utilizes global context information through different-region-based context aggregation. SegNet~\cite{badrinarayanan2017segnet} and U-Net~\cite{ronneberger2015u} used a encoder-decoder architecture for mapping the low-resolution features to input-resolution.

\subsection{Unsupervised Domain Adaptation for Semantic Segmentation}
Unsupervised Domain Adaptation (UDA) is applied to semantic segmentation to save the effort for per-pixel annotation. The goal of the UDA for semantic segmentation is that the model, trained with labeled source data and unlabeled target data, achieves high performance in target data, and this depends on reducing the discrepancy between source and target domains. Many methods, including adversarial learning, are proposed. \citet{du2019ssf} and \citet{wu2018dcan} align latent representations of the two domains in feature space. \citet{hoffman2017cycada} and \citet{zhang2018fully} reduced the visual difference between the two domains for input-level adaptation. \citet{zhu2017unpaired} proposed to transfer the domain of images, and domain-transferred images can be used to learn the segmentation model. \citet{tsai2018learning} and \citet{luo2019taking} use output-level discriminators to adapt the semantic predictions from two domains. \citet{li2019bidirectional} proposed self-training approach to provide extra target information. We deal with the UDA for semantic segmentation by separating content and style from an image and then transferring content from source to target domain.

\subsection{Class Imbalance Problem}
The class imbalance problem occurs when training the model using a dataset with a long-tailed class distribution. This problem stands out in the semantic segmentation task that predicts the pixel-wise category of the image. The urban scene dataset~\cite{cordts2016cityscapes}, which is commonly used for the semantic segmentation, has an imbalanced number of class-wise samples. In this dataset, there are many samples for head classes (e.g., road, sky, etc.), while there are significantly fewer samples for tail classes (e.g., traffic sign, etc.). The model trained with the imbalanced dataset is skewed to the head classes ~\cite{cui2019class}. 

For handling the issue, the class rebalancing approaches have been proposed. The class-balanced loss~\cite{cui2019class} calculates the effective number of samples for each class, and then they rebalance the loss. The cut-and-paste~\cite{dwibedi2017cut}, which cuts and pastes objects into images, can be used to increase the amount of tail-class training data. However, these methods cannot be applied directly to UDA for semantic segmentation because of the absence of the target annotation. In this paper, we propose the content transfer to solve the lack of data for target tail classes, which is the essential cause of the class imbalance problem.

\begin{figure*}[t]
\centering
\includegraphics[width=1.0\textwidth]{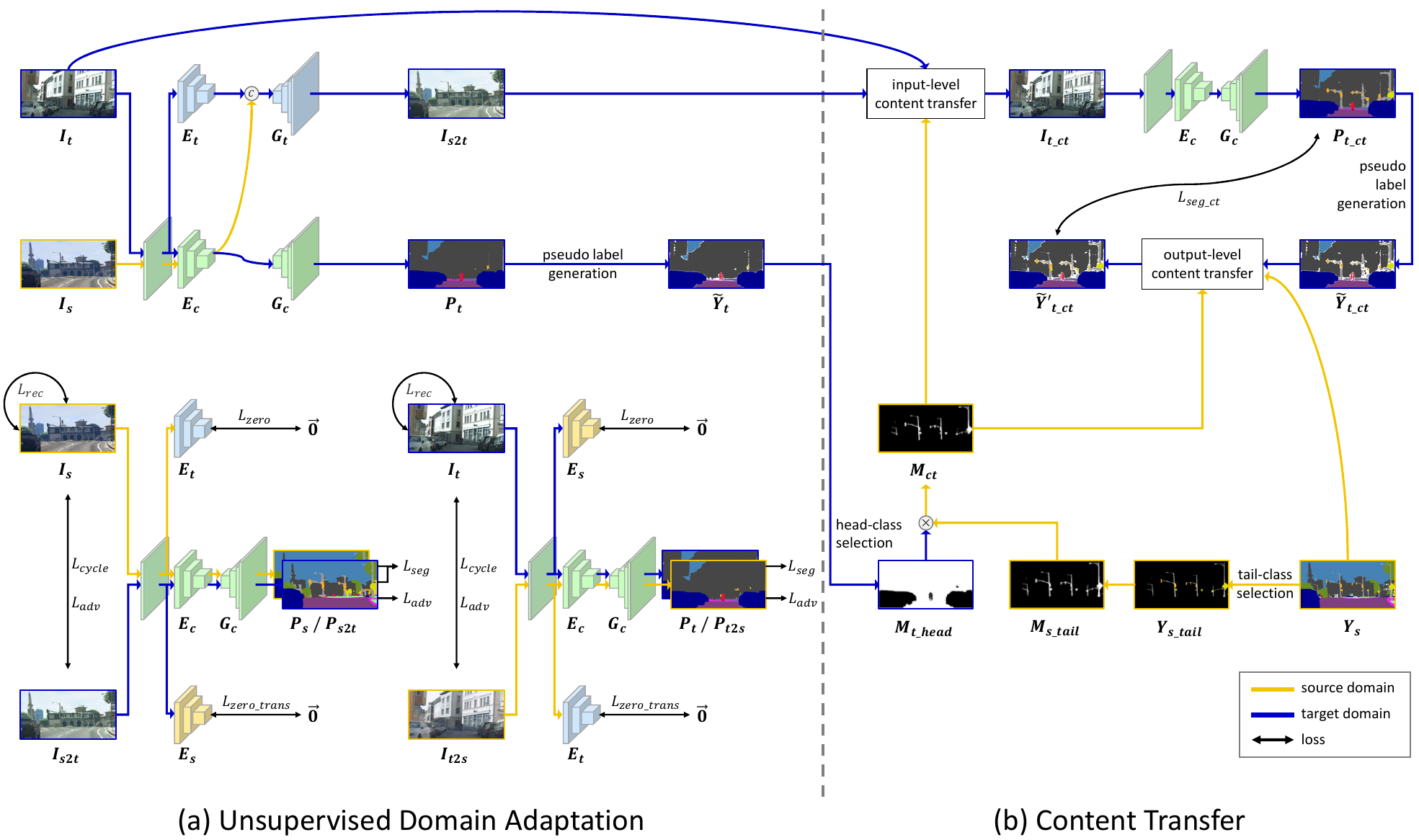}
\caption{An overview of the proposed architecture. Our model consists of two parts, unsupervised domain adaptation (UDA) and content transfer. The UDA part contains domain transfer and semantic segmentation. The image is separated into the domain-specific style and domain-invariant content, and the combination of them enables us to generate the translated image. The content feature is used for both semantic segmentation and content transfer. The content transfer part transmits the tail-class content from source to target in the input and output space. The class imbalance problem is alleviated by learning the segmentation model using the content transferred image and label.}
\label{network}
\end{figure*}

\section{Method}

The overall framework is depicted in Figure ~\ref{network}. We first describe the unsupervised domain adaptation (UDA) part, which contains domain transfer and semantic prediction, with the zero-style loss. Second, we describe the content transfer part to address the class imbalance problem in the UDA for semantic segmentation.

\subsection{Unsupervised Domain Adaptation}

We denote the source domain as $S$ and target domain as $T$. In the UDA, the source image $I_s\in\mathbb{R}^{{H}\times{W}\times{3}}$ with label $Y_s\in\mathbb{R}^{{H}\times{W}\times{K}}$ and target image $I_t\in\mathbb{R}^{{H}\times{W}\times{3}}$ without label are given. The $H$ and $W$ are the height and width of the image and $K$ is the number of the semantic categories. Our model contains one shared content encoder $E_c$, one shared segmenter $G_c$, two style encoders $E_s$ and $E_t$, and two image generators $G_s$ and $G_t$. The $E_c$ captures semantic information regardless of domain. The $E_s$ and $E_t$ capture style information that depends on the source and target domains, respectively. The first convolutional layers of three encoders are shared to extract the features from the raw image before decomposing style and content. The $G_c$ predicts the pixel-level categories of the image using content feature encoded from $E_c$. The $G_s$ and $G_t$ generate style-specific images corresponding to the given style while maintaining the given content. We translate $I_s$ and $I_t$ into the other domain as:
\begin{linenomath}
\begin{align}
&I_{s2t}=G_t(E_t(I_t),E_c(I_s)),\\
&I_{t2s}=G_s(E_s(I_s),E_c(I_t)),
\end{align}
\end{linenomath}
where $I_{s2t}$ and $I_{t2s}$ represent domain-translated images from source to target and from target to source, respectively.

The translated images can be transferred back to the original domain as:
\begin{linenomath}
\begin{align}
    &I_{s2t2s} = G_s(E_s(I_{t2s}),E_c(I_{s2t}))),\\
    &I_{t2s2t} = G_t(E_t(I_{s2t}),E_c(I_{t2s}))),
\end{align}
\end{linenomath}
where $I_{s2t2s}$ represents the image in which domain is transferred to source ${\rightarrow}$ target ${\rightarrow}$ source and $I_{t2s2t}$ represents the image in which domain is transferred to target ${\rightarrow}$ source ${\rightarrow}$ target.

The goal of our domain adaptation is to adapt the contents of the source and target domain, which is the cue for semantic segmentation, not including the styles of both domains at all. To do this, we propose the zero-style loss to ensure that the content encoder captures only the content and the style encoder only captures the style from an image.

\subsubsection{Zero-Style Loss}

Zero loss~\cite{benaim2019domain} has been proposed to create a new image using shared information and unique information extracted independently from two images. We can adapt the zero loss to our model for extracting content and style independently as follows:
\begin{equation}
\begin{aligned}
    L_{zero} = ||E_t(I_s) - 0||_1 + ||E_s(I_t) - 0||_1.\\
\end{aligned}
\end{equation}

The $E_s$ and $E_t$ capture unique information that exists only in the source domain and target domain by $L_{zero}$, so it gets no information from $I_t$ and $I_s$, respectively. However, there is no guarantee that unique information only implies the style of the domain, such as color and texture, and does not include the content for segmentation. The shared content cannot be guaranteed to contain all the required information for semantic segmentation. To address these concerns, we propose the zero-style loss that forces the $E_c$ to extract the content feature for segmentation and induce $E_s$ and $E_t$ to extract only the styles of source and target, respectively. The content features of the two domains are aligned by $L_{seg}$, and this affects the segmentation performance on the target domain. 
\begin{equation}
\begin{aligned}
    L_{seg} = L_{CE}(P_s, Y_s) + L_{CE}(P_{s2t}, Y_{s2t})\\
\end{aligned}
\end{equation}
where $P_s=G_c(E_c(I_s))$ and $P_{s2t}=G_c(E_c(I_{s2t}))$ are output probability maps, and $Y_{s2t}$ is the same as $Y_s$ since the content of $I_{s2t}$ is equal to the content of $I_s$. The $L_{CE}$ denotes a cross-entropy loss function.
\begin{equation}
\begin{aligned}
    L_{CE}(P_s,Y_s) = -\sum_{h}^{H}{\sum_{w}^{W}{\sum_{k}^{K}{Y_s^{(h,w,k)}\mathrm{log}(P_s^{(h,w,k)})}}}.\\
\end{aligned}
\end{equation}

The zero loss also applies to $I_{s2t}$ and $I_{t2s}$ as follows:
\begin{equation}
\begin{aligned}
    L_{zero\_trans} = ||E_s(I_{s2t}) - 0||_1 + ||E_t(I_{t2s}) - 0||_1.\\
\end{aligned}
\end{equation}
The $L_{zero\_trans}$ is used in two perspectives. First, it guarantees the domain-transferred images ($I_{s2t}$ and $I_{t2s}$) are well adapted to each domain. Second, it enables the style encoders ($E_s$ and $E_t$) to capture each style not only from original images ($I_s$ and $I_t$), but also from domain-transferred images ($I_{t2s}$ and $I_{s2t}$).

The proposed zero-style loss enables feature-level alignment in the content feature space, and it can be defined as follows: 
\begin{equation}
\begin{aligned}	
L_{zero-style} = L_{zero}+L_{zero\_trans}+L_{seg}.\\ 
\end{aligned}
\end{equation}

\subsubsection{Cycle-Consistent Adversarial Learning}

In addition to the feature alignment, we apply the adversarial loss $L_{adv}$ to reduce the domain gap in image and prediction~\cite{chang2019all}. Additionally, we define cycle-consistent loss~\cite{zhu2017unpaired} $L_{cycle}$ and reconstruction loss $L_{rec}$ to prevent content information loss that may occur in the process of the domain transfer. 
\begin{equation}
\begin{aligned}
    L_{cycle} = ||I_{s2t2s} - I_s||_1 + ||I_{t2s2t} - I_t||_1,\\
\end{aligned}
\end{equation}
\begin{equation}
\begin{aligned}
    L_{rec} \;&= ||I_{s2s} - I_s||_1+||I_{t2t} - I_t||_1,\;\;\;\;\\
\end{aligned}
\end{equation}
where $I_{s2s}$ and $I_{t2t}$ are reconstructed source and target images, respectively.

\subsubsection{Pseudo Label Generation for Self-Training}

In the UDA settings, there is no label for $I_t$. Our goal is to improve performance on the target, so we train the model with the target pseudo label $\tilde{Y}_t$ assigned by maximum probability threshold (MPT) \cite{li2019bidirectional}. The $\tilde{Y}_t~$ is defined as follows:
\begin{equation}
    \tilde{Y}_{t}^{(h,w,k^{*})} =
    \begin{cases}
    1 & \text{if}\;\;P_{t}^{(h,w,k^*)} > P_{thres}^{k^*}\\
    0 & \text{otherwise}\\
    \end{cases}
\label{Eq_pseudolabel}
\end{equation}
where $P_t$ is the output probability map of $I_t$, and class $k^{*}= \argmax_{k}P_{t}^{(h,w,k)}$ is chosen for each pixel $I_{t}^{(h,w)}$. The class-wise confidence threshold $P_{thres}^{k}$ is newly set for every $P_t$.

\subsection{Content Transfer}

In this section, we introduce the content transfer approach to solve the class imbalance problem in the UDA for semantic segmentation. The fundamental reason for this problem is a lack of samples for tail classes, so we resolve this issue to increase the training samples by transferring the tail-class content from source to target. Our approach is applied to the input and output space.

\subsubsection{Tail-Class Selection}

For the tail-class content transfer, we should select the tail-class region from the source image. We easily obtain the tail-class label $Y_{s\_tail}\in\mathbb{R}^{{H}\times{W}\times{K}}$ as follows:
\begin{equation}
    Y_{s\_tail}^{(h,w,k)} =
    \begin{cases}
    Y_s^{(h,w,k)} & \text{if}\;\;k\;\;\text{is a tail class}\\
    0 & \text{otherwise}\\
    \end{cases}
\end{equation}

\begin{figure}[t]
\centering
\includegraphics[width=0.98\columnwidth]{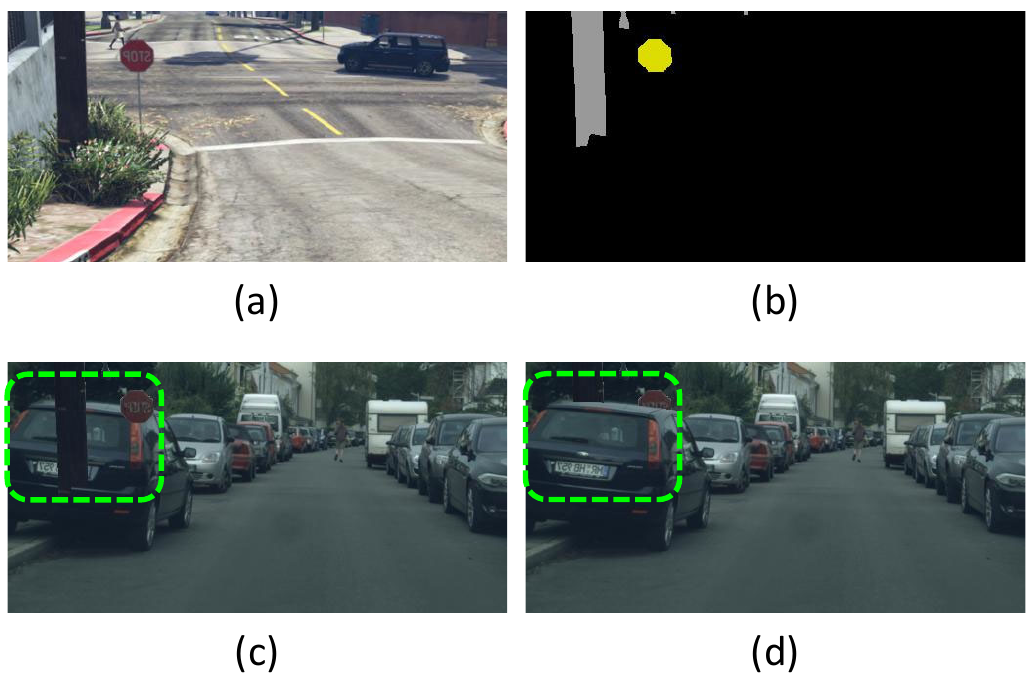}
\caption{The results of the input-level content transfer with and without considering the content of the target image. Given the source image (a), the tail-class region is obtained as (b). For the input-level content transfer, as shown in (c), we can easily transfer all the source content corresponding to all areas of (b) to the target. Since this simple way does not consider the content information of the target image, the original target content is obscured as inside the green dashed box of (c). Therefore, as shown in (d), we transfer the source content only to the head-class region by considering the content information of the target image.}
\label{effect_of_M_ct}
\end{figure}

\begin{figure*}[t]
\centering
\includegraphics[width=0.96\textwidth]{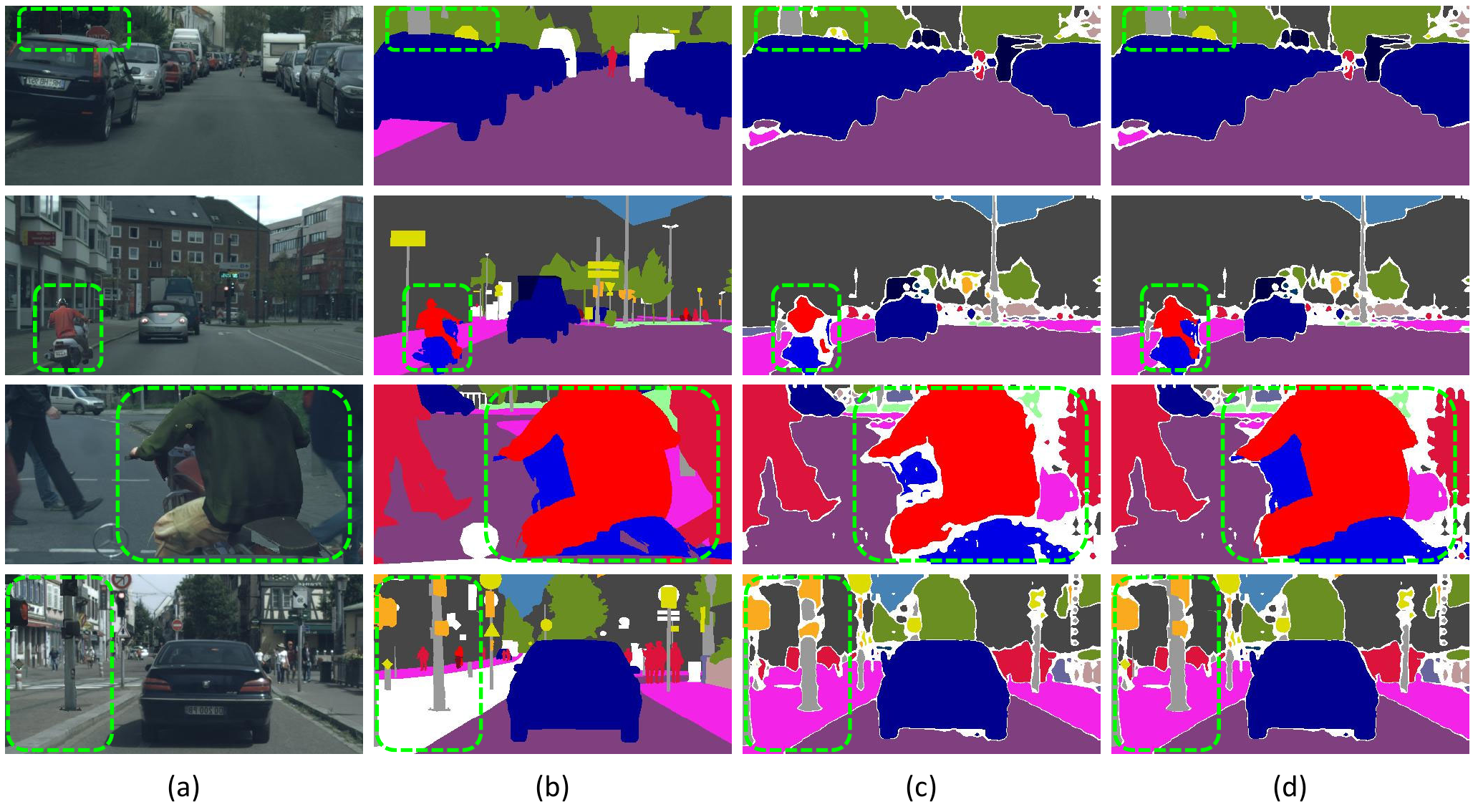}
\caption{The results of the output-level content transfer. (a) is the content-transferred target image, (b) is the ground-truth label, (c) is the pseudo label, and (d) is the improved pseudo label by transferring the source content to the target in output space. The green dashed box indicates the content transmission region.}
\label{output_level_content_transfer}
\end{figure*}

The tail-class content can be directly transferred from source to target by using $Y_{s\_tail}$, but this method does not consider the content information of the target image. Therefore, it happens that the tail-class region of the original target image is occluded by transferred content, as shown in Figure ~\ref{effect_of_M_ct}. To prevent this problem, we refine $Y_{s\_tail}$ to transfer the content only to the head-class region of the original target image. We can obtain the head-class mask from the target image as follows:
\begin{equation}
\begin{aligned}
    M_{t\_head} = {\tilde{Y}_t}{K_{head}}\\
\end{aligned}
\end{equation}
where $K_{head}$ is a K-dimensional vector with a head-class of 1 and a tail-class of 0. Then, we define a mask $M_{ct}$ for the content transfer.
\begin{equation}
\begin{aligned}
    M_{ct} = {M_{t\_head}}\odot{M_{s\_tail}}\\
\end{aligned}
\end{equation}
where $M_{s\_tail}$ is a binary mask of $Y_{s\_tail}$.

\subsubsection{Input-Level Content Transfer}

In the input-level content transfer, we transmit the tail-class content from domain-translated image $I_{s2t}$ to target image $I_t$. It can be formulated as follows:  
\begin{equation}
\label{Eq_inct}
\begin{aligned}
 I_{t\_ct}={M_{ct}}\odot{I_{s2t}} + {(1-{M_{ct}})}\odot{I_t}.\\
\end{aligned}
\end{equation}

As shown in Figure ~\ref{effect_of_M_ct}, $I_{t\_ct}$ is created in consideration of the target content. We predict the semantic categories for the content-transferred image $I_{t\_ct}$, and the output probability map $P_{t\_ct}$ of $I_{t\_ct}$ is obtained by $G_c(E_c(I_{t\_ct}))$. We generate a pseudo label $\tilde{Y}_{t\_ct}$ for $I_{t\_ct}$ using Equation~(\ref{Eq_pseudolabel}), and train the model with the generated data. This approach solves the class imbalance issue by increasing the tail-class samples for training the network.

\subsubsection{Output-Level Content Transfer}

The semantic segmentation model biased toward head classes produces the probability map with low confidence for tail classes, so the pseudo label $\tilde{Y}_{t\_ct}$ is highly incomplete for tail classes. To enable more reliable supervision for tail-class content, we also transfer the source content to the target in the output space. Based on the fact that the label for the transferred region of $I_{t\_ct}$ is the same as the $Y_s$, we can improve the quality of the pseudo label for $I_{t\_ct}$ as follows:
\begin{equation}
\label{Eq_outct}
\begin{aligned}
 \tilde{Y}_{t\_ct}'= {M_{ct}}\odot{Y_{s}} + {(1-{M_{ct}})}\odot{\tilde{Y}_{t\_ct}}.\\
\end{aligned}
\end{equation}

The pseudo label with the output-level content transfer is shown in Figure~\ref{output_level_content_transfer}. We can train the segmentation model with the definite label for the transferred region. The segmentation loss for the content transferred data is defined as follows:
\begin{equation}
\begin{aligned}
    L_{seg\_ct} \;&= \,L_{CE}(P_{t\_ct}, \tilde{Y}_{t\_ct}').\\
\end{aligned}
\end{equation}

\subsection{Overall Loss Function}

Our model is based on the DISE~\cite{chang2019all} and inserted the proposed approaches. The overall loss function is as follows:
\begin{equation}
\begin{aligned}
    L_{total} = L_{DISE} + L_{zero-style} + L_{cycle} + L_{seg\_ct},\\
\end{aligned}
\end{equation}
where $L_{DISE}$ is the loss function of DISE and contains $L_{rec}$ and $L_{adv}$.

\section{Experiments}\label{sec:Experiments}

\begin{table*}[t]
\centering
\caption{Results of adapting GTA5 to Cityscapes. All methods use DeepLab v2 with the backbone ResNet-101 as the segmentation network. We measure the mIoU performance of the 19 classes in the evaluation set of Cityscapes.}
\label{table:GTA5}
\resizebox{\textwidth}{!}{\begin{tabular}[b]{l|ccccccccccccccccccc|c}\hline\noalign{\smallskip}
\multicolumn{21}{c}{GTA5 $\rightarrow$ Cityscapes}\\\noalign{\smallskip}\hline
Methods            & {\rotatebox{90}{road }}          & {\rotatebox{90}{sidewalk }}      & {\rotatebox{90}{building }}      & {\rotatebox{90}{wall }}          & {\rotatebox{90}{fence }}         & {\rotatebox{90}{pole }}          & {\rotatebox{90}{traffic light }} & {\rotatebox{90}{traffic sign }}  & {\rotatebox{90}{vegetation }}    & {\rotatebox{90}{terrain }}       & {\rotatebox{90}{sky }}           & {\rotatebox{90}{person }}        & {\rotatebox{90}{rider }}         & {\rotatebox{90}{car }}           & {\rotatebox{90}{truck }}         & {\rotatebox{90}{bus }}           & {\rotatebox{90}{train }}        & {\rotatebox{90}{motorcycle }}    & {\rotatebox{90}{bicycle }}       & mIoU          \\\hline
NonAdapt                                 & 75.8 & 16.8 & 77.2 & 12.5 & 21.0 & 25.5 & 30.1 & 20.1 & 81.3 & 24.6 & 70.3 & 53.8 & 26.4 & 49.9 & 17.2 & 25.9 & 6.5 & 25.3 & 36.0 & 36.6 \\
AdaptSegNet \cite{tsai2018learning}      & 86.5 & 36.0 & 79.9 & 23.4 & 23.3 & 23.9 & 35.2 & 14.8 & 83.4 & 33.3 & 75.6 & 58.5 & 27.6 & 73.7 & 32.5 & 35.4 & 3.9 & 30.1 & 28.1 & 42.4 \\
CLAN \cite{luo2019taking}                & 87.0 & 27.1 & 79.6 & 27.3 & 23.3 & 28.3 & 35.5 & 24.2 & 83.6 & 27.4 & 74.2 & 58.6 & 28.0 & 76.2 & 33.1 & 36.7 & 6.7 & 31.9 & 31.4 & 43.2 \\
MaxSquare \cite{chen2019domain}          & 88.1 & 27.7 & 80.8 & 28.7 & 19.8 & 24.9 & 34.0 & 17.8 & 83.6 & 34.7 & 76.0 & 58.6 & 28.6 & 84.1 & 37.8 & 43.1 & 7.2 & 32.2 & 34.2 & 44.3 \\
SSF-DAN \cite{du2019ssf}                 & 90.3 & 38.9 & 81.7 & 24.8 & 22.9 & 30.5 & 37.0 & 21.2 & 84.8 & 38.8 & 76.9 & 58.8 & 30.7 & 85.7 & 30.6 & 38.1 & 5.9 & 28.3 & 36.9 & 45.4 \\
DISE \cite{chang2019all}                 & 91.5 & 47.5 & 82.5 & 31.3 & 25.6 & \textbf{33.0} & 33.7 & 25.8 & 82.7 & 28.8 & 82.7 & 62.4 & 30.8 & 85.2 & 27.7 & 34.5 & 6.4 & 25.2 & 24.4 & 45.4 \\
AdvEnt+MinEnt \cite{vu2019advent}        & 89.4 & 33.1 & 81.0 & 26.6 & 26.8 & 27.2 & 33.5 & 24.7 & 83.9 & 36.7 & 78.8 & 58.7 & 30.5 & 84.8 & \textbf{38.5} & 44.5 & 1.7 & 31.6 & 32.4 & 45.5 \\
APODA \cite{yang2020adversarial}         & 85.6 & 32.8 & 79.0 & 29.5 & 25.5 & 26.8 & 34.6 & 19.9 & 83.7 & 40.6 & 77.9 & 59.2 & 28.3 & 84.6 & 34.6 & 49.2 & 8.0 & 32.6 & 39.6 & 45.9 \\
MaxSquare+IW+Multi \cite{chen2019domain} & 89.4 & 43.0 & 82.1 & 30.5 & 21.3 & 30.3 & 34.7 & 24.0 & 85.3 & 39.4 & 78.2 & \textbf{63.0} & 22.9 & 84.6 & 36.4 & 43.0 & 5.5 & \textbf{34.7} & 33.5 & 46.4 \\
Patch Alignment \cite{tsai2019domain}    & 92.3 & 51.9 & 82.1 & 29.2 & 25.1 & 24.5 & 33.8 & 33.0 & 82.4 & 32.8 & 82.2 & 58.6 & 27.2 & 84.3 & 33.4 & 46.3 & 2.2 & 29.5 & 32.3 & 46.5 \\
WeakSegDA \cite{paul2020domain}          & 91.6 & 47.4 & 84.0 & 30.4 & \textbf{28.3} & 31.4 & \textbf{37.4} & 35.4 & 83.9 & 38.3 & 83.9 & 61.2 & 28.2 & 83.7 & 28.8 & 41.3 & 8.8 & 24.7 & \textbf{46.4} & 48.2 \\
BDL \cite{li2019bidirectional}           & 91.0 & 44.7 & 84.2 & \textbf{34.6} & 27.6 & 30.2 & 36.0 & 36.0 & 85.0 & \textbf{43.6} & 83.0 & 58.6 & \textbf{31.6} & 83.3 & 35.3 & \textbf{49.7} & 3.3 & 28.8 & 35.6 & 48.5\\
\textbf{Ours}                            & \textbf{95.3} & \textbf{65.1} & \textbf{84.6} & 33.2 & 23.7 & 32.8 & 32.7 & \textbf{36.9} & \textbf{86.0} & 41.0 & \textbf{85.6} & 56.1 & 25.9 & \textbf{86.3} & 34.5 & 39.1 & \textbf{11.5} & 28.3 & 43.0 & \textbf{49.6} \\\hline
\end{tabular}}
\end{table*}

\begin{table*}[t]
\centering
\caption{Results of adapting SYNTHIA to Cityscapes. All methods use DeepLab v2 with the backbone ResNet-101 as the segmentation network. We measure the mIoU of the 16 classes in the evaluation set of Cityscapes and the mIoU of the 13 classes (mIoU*), excluding classes with *.}
\label{table:SYNTHIA}
\resizebox{\textwidth}{!}{\begin{tabular}[b]{l|cccccccccccccccc|cc}\hline\noalign{\smallskip}
\multicolumn{19}{c}{SYNTHIA $\rightarrow$ Cityscapes}\\\noalign{\smallskip}\hline
Methods            & {\rotatebox{90}{road }}          & {\rotatebox{90}{sidewalk }}      & {\rotatebox{90}{building }}      & {\rotatebox{90}{wall* }}          & {\rotatebox{90}{fence* }}         & {\rotatebox{90}{pole* }}          & {\rotatebox{90}{traffic light }} & {\rotatebox{90}{traffic sign }}  & {\rotatebox{90}{vegetation }}    & {\rotatebox{90}{sky }}           & {\rotatebox{90}{person }}        & {\rotatebox{90}{rider }}         & {\rotatebox{90}{car }}           & {\rotatebox{90}{bus }}           & {\rotatebox{90}{motorcycle }}    & {\rotatebox{90}{bicycle }}       & mIoU                                      & mIoU*\\\hline
NonAdapt                                 & 55.6 & 23.8 & 74.6 & 9.2  & 0.2 & 24.4 & 6.1  & 12.1 & 74.8 & 79.0 & 55.3 & 19.1 & 39.6 & 23.3 & 13.7 & 25.0 & 33.5 & 38.6\\
MaxSquare \cite{chen2019domain}          & 77.4 & 34.0 & 78.7 & 5.6  & 0.2 & 27.7 & 5.8  & 9.8  & 80.7 & 83.2 & 58.5 & 20.5 & 74.1 & 32.1 & 11.0 & 29.9 & 39.3 & 45.8\\
Patch Alignment \cite{tsai2019domain}    & 82.4 & 38.0 & 78.6 & 8.7  & 0.6 & 26.0 & 3.9  & 11.1 & 75.5 & 84.6 & 53.5 & 21.6 & 71.4 & 32.6 & 19.3 & 31.7 & 40.0 & 46.5\\
AdaptSegNet \cite{tsai2018learning}      & 84.3 & 42.7 & 77.5 & -    & -   & -    & 4.7  & 7.0  & 77.9 & 82.5 & 54.3 & 21.0 & 72.3 & 32.2 & 18.9 & 32.3 & -    & 46.7\\
CLAN \cite{luo2019taking}                & 81.3 & 37.0 & 80.1 & -    & -   & -    & 16.1 & 13.7 & 78.2 & 81.5 & 53.4 & 21.2 & 73.0 & 32.9 & 22.6 & 30.7 & -    & 47.8\\
AdvEnt+MinEnt \cite{vu2019advent}        & 85.6 & 42.2 & 79.7 & 8.7  & 0.4 & 25.9 & 5.4  & 8.1  & 80.4 & 84.1 & 57.9 & 23.8 & 73.3 & 36.4 & 14.2 & 33.0 & 41.2 & 48.0\\
MaxSquare+IW+Multi \cite{chen2019domain} & 82.9 & 40.7 & 80.3 & 10.2 & \textbf{0.8} & 25.8 & 12.8 & 18.2 & 82.5 & 82.2 & 53.1 & 18.0 & 79.0 & 31.4 & 10.4 & 35.6 & 41.4 & 48.2\\
DISE \cite{chang2019all}                 & 91.7 & 53.5 & 77.1 & 2.5  & 0.2 & 27.1 & 6.2  & 7.6  & 78.4 & 81.2 & 55.8 & 19.2 & 82.3 & 30.3 & 17.1 & 34.3 & 41.5 & 48.8\\
SSF-DAN \cite{du2019ssf}                 & 84.6 & 41.7 & 80.8 & -    & -   & -    & 11.5 & 14.7 & 80.8 & \textbf{85.3} & 57.5 & 21.6 & 82.0 & 36.0 & 19.3 & 34.5 & -    & 50.0\\
BDL \cite{li2019bidirectional}           & 86.0 & 46.7 & 80.3 & - & - & - & 14.1 & 11.6 & 79.2 & 81.3 & 54.1 & \textbf{27.9} & 73.7 & 42.2 & 25.7 & 45.3 & -    & 51.4\\
WeakSegDA \cite{paul2020domain}          & 92.0 & 53.5 & 80.9 & 11.4 & 0.4 & 21.8 & 3.8 & 6.0 & 81.6 & 84.4 & \textbf{60.8} & 24.4 & 80.5 & 39.0 & \textbf{26.0} & 41.7 & 44.3 & 51.9\\
APODA \cite{yang2020adversarial}         & 86.4 & 41.3 & 79.3 & -    & -    & -    & \textbf{22.6} & 17.3 & 80.3 & 81.6 & 56.9 & 21.0 & \textbf{84.1} & \textbf{49.1} & 24.6 & 45.7 & -    &53.1\\
\textbf{Ours}                            & \textbf{93.3} & \textbf{54.0} & \textbf{81.3} & \textbf{14.3}  & 0.7 & \textbf{28.8} & 21.3 & \textbf{22.8} & \textbf{82.6} & 83.3 & 57.7 & 22.8 & 83.4 & 30.7 & 20.2 & \textbf{47.2} & \textbf{46.5} & \textbf{53.9}\\\hline
\end{tabular}}
\end{table*}

\subsection{Datasets}

In this works, we consider two common synthetic-to-real domain adaptation problems; GTA5\cite{richter2016playing} $\rightarrow$ Cityscapes\cite{cordts2016cityscapes} and SYNTHIA\cite{ros2016synthia} $\rightarrow$ Cityscapes. We use the synthetic dataset (GTA5 or SYNTHIA) with pixel-level semantic annotations as the source domain and real-world dataset (Cityscapes) without annotation as the target domain during training. For both synthetic-to-real experiments, we evaluate the segmentation model in the validation set of the Cityscapes.\\
\textbf{GTA5} is the large-scale synthetic dataset containing $24966$ urban scene images with a resolution of $1914\times1052$. These high-resolution images are rendered from Grand Theft Auto V game engine and automatically per-pixel annotated into $19$ semantic categories. We consider all $19$ classes, which are fully compatible with Cityscapes.\\
\textbf{SYNTHIA} is another synthetic dataset rendered from a virtual city. We use the SYNTHIA-RAND-CITYSCAPES subset, which contains 9400 labeled images with a resolution of $1280\times760$ and $16$ common categories with Cityscapes. We consider both $16$ and $13$ classes, following the standard protocol.\\
\textbf{Cityscapes} is the real-world dataset that contains $5000$ urban street scenes. The dataset consists of a training set with $2975$ images, a validation set with $500$ images, and a test set with $1525$ images. The images have a resolution of $2048\times1024$ and are annotated into $19$ classes. We use the training set without labels during the learning process and evaluate the segmentation model in the validation set.

\begin{table*}[t]
\centering
\caption{Ablation studies of adapting GTA5 to Cityscapes.}
\label{table:GTA5_Ablation}
\resizebox{\textwidth}{!}{\begin{tabular}[b]{l|ccccc|ccccccccccccccccccc|cl}\hline\noalign{\smallskip}
\multicolumn{27}{c}{GTA5 $\rightarrow$ Cityscapes}\\\noalign{\smallskip}\hline
\multirow[b]{5}{*}{Methods}& \multicolumn{3}{c}{\multirow[b]{4}{*}{UDA}} & \multicolumn{2}{c|}{\multirow[b]{4}{*}{CT}} & \multirow[b]{5}{*}{\rotatebox{90}{road }}          & \multirow[b]{5}{*}{\rotatebox{90}{sidewalk }}      & \multirow[b]{5}{*}{\rotatebox{90}{building }}      & \multirow[b]{5}{*}{\rotatebox{90}{wall }}          & \multirow[b]{5}{*}{\rotatebox{90}{fence }}         & \multirow[b]{5}{*}{\rotatebox{90}{pole }}          & \multirow[b]{5}{*}{\rotatebox{90}{traffic light }} & \multirow[b]{5}{*}{\rotatebox{90}{traffic sign }}  & \multirow[b]{5}{*}{\rotatebox{90}{vegetation }}    & \multirow[b]{5}{*}{\rotatebox{90}{terrain }}       & \multirow[b]{5}{*}{\rotatebox{90}{sky }}           & \multirow[b]{5}{*}{\rotatebox{90}{person }}        & \multirow[b]{5}{*}{\rotatebox{90}{rider }}         & \multirow[b]{5}{*}{\rotatebox{90}{car }}           & \multirow[b]{5}{*}{\rotatebox{90}{truck }}         & \multirow[b]{5}{*}{\rotatebox{90}{bus }}           & \multirow[b]{5}{*}{\rotatebox{90}{train }}        & \multirow[b]{5}{*}{\rotatebox{90}{motorcycle }}    & \multirow[b]{5}{*}{\rotatebox{90}{bicycle }} & \multirow[b]{5}{*}{mIoU} & \multirow[b]{5}{*}{mIoU$^\text{tail}$} \\
 & \multicolumn{3}{c}{} & \multicolumn{2}{c|}{} & & & & & & & & & & & & & & & & & & & &  \\
& & & & & & & & & & & & & & & & & & & & & & & & & & \\
& & & & & & & & & & & & & & & & & & & & & & & & & & \\
     & Sty & Cyc & ST & In & Out & & & & & & & & & & & & & & & & & & & & & \\\hline
DISE &     &     &    &    &     & 91.5 & 47.5 & 82.5 & 31.3 & 25.6 & \textbf{33.0} & \textbf{33.7} & 25.8 & 82.7 & 28.8 & 82.7 & \textbf{62.4} & \textbf{30.8} & 85.2 & 27.7 & 34.5 & 6.4 & 25.2 & 24.4 & 45.4 & 28.8 \\\hline
\multirow{5}{*}{Ours} & \checkmark                &                           &                           &                           &                           & 91.6 & 51.2 & 82.7 & 33.0 & 25.9 & 29.3 & 32.7 & 29.9 & 82.4 & 25.6 & 81.7 & 59.6 & 29.6 & 85.1 & 30.9 & 43.5 & 6.1 & 29.1 & 34.8 & 46.6                 & 30.9 (+7.3\%)                   \\
                      & \checkmark                &\checkmark                 &                           &                           &                           & 91.9 & 51.9 & 82.8 & 34.1 & \textbf{26.2} & 29.3 & 33.1 & 31.6 & 82.9 & 28.2 & 81.0 & 59.5 & 29.1 & 85.7 & 31.2 & \textbf{43.9} & 8.0  & \textbf{28.4} & 32.4 & 46.9                 & 30.6 (+6.3\%)                   \\
                      &\checkmark                 &\checkmark                 &\checkmark                 &                           &                           & 93.9 & 61.5 & 83.8 & \textbf{36.8} & 23.7 & 28.1 & 30.9 & 32.4 & 85.2 & 35.5 & 83.0 & 52.6 & 26.6 & 86.5 & 33.2 & 43.7 & 12.6 & 25.1 & 36.4 & 48.0                 & 29.9 (+3.8\%)                   \\
                      &\checkmark                 &\checkmark                 &\checkmark                 &\checkmark                 &                           & 94.5 & 62.6 & 84.3 & 35.9 & 24.4 & 30.2 & 30.2 & 34.9 & 85.3 & 35.7 & 84.0 & 55.6 & 27.7 & \textbf{86.7} & 31.8 & 43.6 & \textbf{12.7} & 26.7 & 39.2 & 48.7                 & 31.5 (+9.4\%)                     \\
                      &\checkmark                 &\checkmark                 &\checkmark                 &\checkmark                 &\checkmark                 & \textbf{95.3} & \textbf{65.1} & \textbf{84.6} & 33.2 & 23.7 & 32.8 & 32.7 & \textbf{36.9} & \textbf{86.0} & \textbf{41.0} & \textbf{85.6} & 56.1 & 25.9 & 86.3 & \textbf{34.5} & 39.1 & 11.5 & 28.3 & \textbf{43.0} & \textbf{49.6}                 & \textbf{33.3 (+15.6\%)}                  \\\hline
\end{tabular}}
\end{table*}

\begin{table*}[t]
\centering
\caption{Ablation studies of adapting SYNTHIA to Cityscapes.}
\label{table:SYNTHIA_Ablation}
\resizebox{\textwidth}{!}{\begin{tabular}[b]{l|ccccc|cccccccccccccccc|cl}\hline\noalign{\smallskip}
\multicolumn{24}{c}{SYNTHIA $\rightarrow$ Cityscapes}\\\noalign{\smallskip}\hline
\multirow[b]{5}{*}{Methods}& \multicolumn{3}{c}{\multirow[b]{4}{*}{UDA}} & \multicolumn{2}{c|}{\multirow[b]{4}{*}{CT}} & \multirow[b]{5}{*}{\rotatebox{90}{road }}          & \multirow[b]{5}{*}{\rotatebox{90}{sidewalk }}      & \multirow[b]{5}{*}{\rotatebox{90}{building }}      & \multirow[b]{5}{*}{\rotatebox{90}{wall }}          & \multirow[b]{5}{*}{\rotatebox{90}{fence }}         & \multirow[b]{5}{*}{\rotatebox{90}{pole }}          & \multirow[b]{5}{*}{\rotatebox{90}{traffic light }} & \multirow[b]{5}{*}{\rotatebox{90}{traffic sign }}  & \multirow[b]{5}{*}{\rotatebox{90}{vegetation }}    & \multirow[b]{5}{*}{\rotatebox{90}{sky }}           & \multirow[b]{5}{*}{\rotatebox{90}{person }}        & \multirow[b]{5}{*}{\rotatebox{90}{rider }}         & \multirow[b]{5}{*}{\rotatebox{90}{car }}           & \multirow[b]{5}{*}{\rotatebox{90}{bus }}           & \multirow[b]{5}{*}{\rotatebox{90}{motorcycle }}    & \multirow[b]{5}{*}{\rotatebox{90}{bicycle }} & \multirow[b]{5}{*}{mIoU} & \multirow[b]{5}{*}{mIoU$^\text{tail}$} \\
 & \multicolumn{3}{c}{} & \multicolumn{2}{c|}{} & & & & & & & & & & & & & & & & &  \\
& & & & & & & & & & & & & & & & & & & & & & & \\
& & & & & & & & & & & & & & & & & & & & & & & \\
     & Sty & Cyc & ST & In & Out & & & & & & & & & & & & & & & & & & \\\hline
DISE &     &     &    &    &     & 91.7 & 53.5 & 77.1 & 2.5  & 0.2 & 27.1 & 6.2  & 7.6  & 78.4 & 81.2 & 55.8 & 19.2 & 82.3 & 30.3 & 17.1 & 34.3 & 41.5 & 18.6 \\\hline
\multirow{5}{*}{Ours} & \checkmark                &                           &                           &                           &                           & 91.3 & 52.4 & 78.6 & 7.1  & 1.3 & 26.7 & 19.6 & 17.5 & 78.4 & 78.3 & 53.1 & 20.1 & 81.1 & 29.0 & 16.0 & 42.4 & 43.3 & 23.7 (+27.4\%)                   \\
                      & \checkmark                &\checkmark                 &                           &                           &                           & 91.6 & 53.0 & 78.9 & 6.1  & 1.2 & 26.8 & 19.1 & 17.7 & 78.5 & 78.4 & 53.5 & 20.2 & 81.5 & 29.3 & 16.4 & 43.4 & 43.5 & 24.0 (+29.0\%)                   \\
                      &\checkmark                 &\checkmark                 &\checkmark                 &                           &                           & 91.7 & 52.3 & 78.8 & 13.7 & \textbf{2.2} & 26.2 & 13.8 & 18.3 & 79.8 & 79.5 & 53.8 & 21.0 & 81.9 & 29.8 & 14.3 & 45.8 & 43.9 & 23.2 (+24.7\%)                   \\
                      &\checkmark                 &\checkmark                 &\checkmark                 &\checkmark                 &                           & \textbf{93.3} & \textbf{55.2} & 81.2 & \textbf{14.7} & 0.8 & 27.9 & 20.1 & 21.8 & 82.5 & 83.1 & 57.2 & \textbf{23.9} & 83.2 & 30.2 & 17.7 & 46.3 & 46.2 & 26.3 (+41.4\%)                     \\
                      &\checkmark                 &\checkmark                 &\checkmark                 &\checkmark                 &\checkmark                 & \textbf{93.3} & 54.0 & \textbf{81.3} & 14.3 & 0.7 & \textbf{28.8} & \textbf{21.3} & \textbf{22.8} & \textbf{82.6} & \textbf{83.3} & \textbf{57.7} & 22.8 & \textbf{83.4} & \textbf{30.7} & \textbf{20.2} & \textbf{47.2} & \textbf{46.5} & \textbf{27.2 (+46.2\%)}                  \\\hline
\end{tabular}}
\end{table*}

\subsection{Implementation Details}\label{sec:implementation}

We implement with PyTorch~\cite{paszke2017automatic}, and all experiments are conducted using a single Nvidia Titan RTX GPU and an Intel Core i7-9700K CPU. Our model is based on the DISE~\cite{chang2019all}, which includes three encoders, two decoders (one segmenter and one generator), and discriminators. For cycle-consistency learning, we use two generators for the source and target domains differently from DISE. The segmentation model is based on DeepLab v2 with the backbone ResNet-101, and the output of the first convolutional layer is shared to capture the content and style features. For adversarial learning, discriminators and three pairs of encoder and decoder are trained alternately. Discriminators are learned to determine whether the pixel-level semantic prediction comes from the source domain or the target domain and to determine whether the generated source and target style images are real or fake. Generators are trained to generate the reconstructed images and style translated images, and the segmenter is trained with the class balanced weight factor~\cite{cui2019class} for source data to predict the pixel-level semantic probabilities. In the self-training, we use the maximum probability threshold~\cite{li2019bidirectional} to assign the pseudo label of the target image. For every target probability map, we sort the class-wise probabilities and set the probability corresponding to the top $50\%$ as the class-wise confidence threshold. If the threshold is greater than $0.9$, we set it to $0.9$.

In this paper, we select the pole, traffic light, traffic sign, rider, motorcycle, and bicycle as tail classes. We selectively use the tail-class samples for the content transfer. The pole is used for content transfer when it exists together with the traffic light or traffic sign, and the rider is used for content transfer when it exists together with the motorcycle or bicycle. We pre-calculate the median value for the number of tail-class samples per image in the source dataset and perform the content transfer during learning when the number of tail-class samples in the source image is higher than that value.

During the training, we resize the input image to a size of $1024\times512$ and then randomly crop to a $512\times256$ size patch. We first train the UDA part to generate a reliable target pseudo label, and then we add the content transfer part and train together. This process reduces the content transfer failures that may occur due to the inaccuracy of the target pseudo label. We train the model with a batch size $2$ and use the weights provided in \cite{chang2019all} as initial weights. The batch normalization layers are freeze. We use the SGD solver with an initial learning rate of ${2.5}\times{10^{-6}}$ and a momentum of $0.9$ for the segmentation model and the Adam solver with an initial learning rate of ${1.0}\times{10^{-5}}$  and betas ${\beta}_1=0.5$, ${\beta}_2=0.999$ for the generators. We use the Adam solver with an initial learning rate of ${1.0}\times{10^{-6}}$ to optimize discriminators for segmentation and image, and they have betas ${\beta}_1=0.9$, ${\beta}_2=0.99$ and ${\beta}_1=0.5$, ${\beta}_2=0.999$, respectively. All the learning rates decrease according to the polynomial policy. For a fair comparison with DISE at the evaluation time, we use a $1024\times512$ sized input to obtain the output prediction of $1024\times512$ size and then resize it to $2048\times1024$.

\subsection{Experimental Results}

We compare our method with existing state-of-the-art UDA methods. We report the results for GTA5$\rightarrow$Cityscapes in Table~\ref{table:GTA5} and the results for SYNTHIA$\rightarrow$Cityscapes in Table~\ref{table:SYNTHIA}. All methods are trained using the labeled GTA5 or SYNTHIA as the source and the unlabeled Cityscapes training set as the target, and they are based on DeepLab v2 with the backbone ResNet-101. We use the mean of class-wise Intersection-over-Union (mIOU) as an evaluation metric, and we measure the performance on the validation set of Cityscapes. All UDA methods have higher performance than “NonAdapt,” which learns the segmentation model using only the source dataset, and it suggests that domain adaptation is effective in semantic segmentation problem. We aim at not only improving the performance of each tail-class but also preventing performance degradation in head classes. Thus our method keeps a balance between head and tail classes, and we achieve the state-of-the-art performance.

\subsection{Ablation Study}
Tables~\ref{table:GTA5_Ablation} and ~\ref{table:SYNTHIA_Ablation} show the effect of adding each component of the proposed method to the base network. “UDA” and “CT” represent the unsupervised domain adaptation part and content transfer part in our method, respectively. “Sty” means training with the zero-style loss, and “Cyc” means cycle-consistency learning, and “ST” means self-training with the pseudo label of the target image. “In” refers to the input-level content transfer, and “Out” refers to the output-level content transfer. “mIoU$^\text{tail}$” is the mIoU of the tail classes. 

As shown in Tables~\ref{table:GTA5_Ablation} and \ref{table:SYNTHIA_Ablation}, we increase the performance by training the model with “Sty” and “Cyc.” The “ST” improves mIoU (46.9 $\rightarrow$ 48.0 and 43.5 $\rightarrow$ 43.9), but it rather degrades mIoU$^\text{tail}$ (30.6 $\rightarrow$ 29.9 and 24.0 $\rightarrow$ 23.2). It means that self-training using a class-imbalanced dataset can make the class imbalance problem worse. That is the point that content transfer exploits in UDA. Even when “InCT” is applied alone, mIoU$^\text{tail}$ is significantly improved (29.9 $\rightarrow$ 31.5 and 23.2 $\rightarrow$ 26.3), and mIoU is also enhances (48.0 $\rightarrow$ 48.7 and 43.9 $\rightarrow$ 46.2). When “OutCT” is also applied, both mIoU$^\text{tail}$ (31.5 $\rightarrow$ 33.3 and 26.3 $\rightarrow$ 27.2) and mIoU (48.7 $\rightarrow$ 49.6 and 46.2 $\rightarrow$ 46.5) are improved from “InCT”. Here, the improvement from “InCT” to “OutCT” on SYNTHIA $\rightarrow$ Cityscapes (+ 0.3) is lower than GTA5 $\rightarrow$ Cityscapes (+~0.9). It demonstrates the limitation of our “OutCT”. When a content taken from the source data is transferred to the target data, the content is placed at the same location in the target data by Equations~(\ref{Eq_inct}) and (\ref{Eq_outct}). However, the vertical angle between SYNTHIA and Cityscapes is quite different, so some objects can be transferred in the unlikely region in Cityscapes (e.g., a rider in the sky), degrading “OutCT.” It can be improved with a context-aware content transfer, and we leave this for future works. 

\begin{figure*}[t]
\centering
\includegraphics[width=1.0\textwidth]{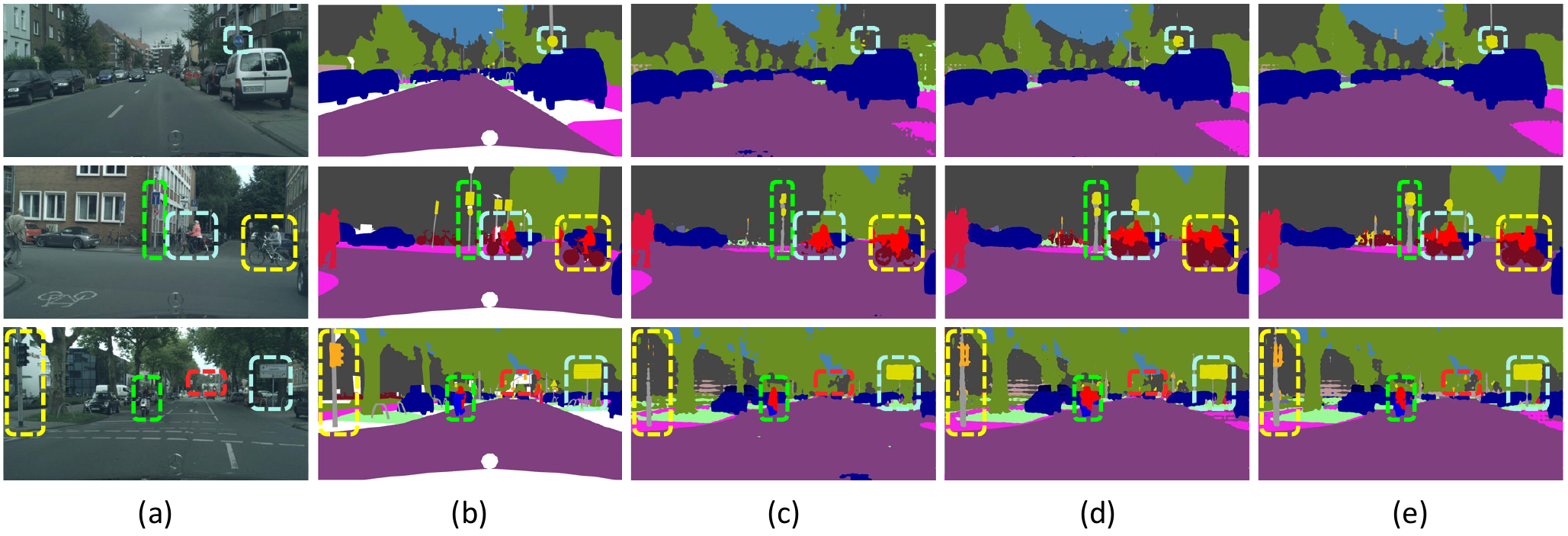}
\caption{Qualitative segmentation results of UDA for GTA5 $\rightarrow$ Cityscapes. (a) Original inputs of Cityscapes; (b) Ground truths; Prediction results from (c) DISE, (d) Ours without OutCT, and (e) Ours.}
\label{result_cityval_using_gta5}
\end{figure*}

\begin{figure*}
\centering
\includegraphics[width=1.0\textwidth]{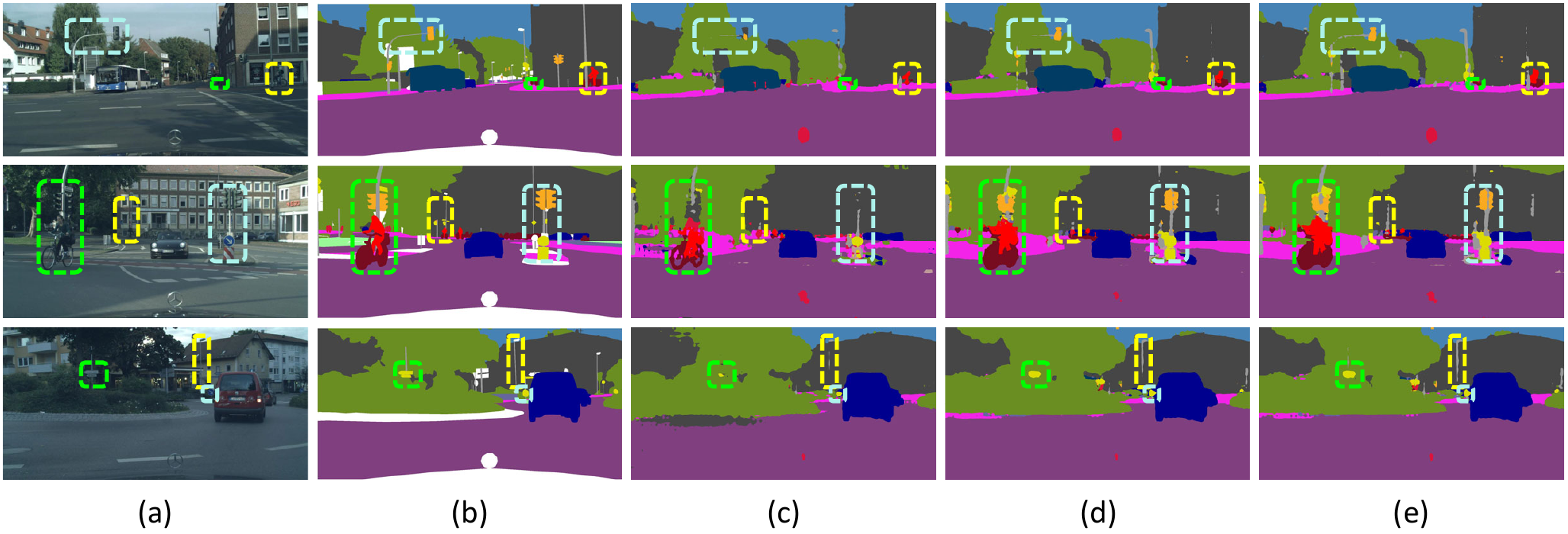}
\caption{Qualitative segmentation results of UDA for SYNTHIA $\rightarrow$ Cityscapes. (a) Original inputs of Cityscapes; (b) Ground truths; Prediction results from (c) DISE, (d) Ours without OutCT, and (e) Ours.}
\label{result_cityval_using_synthia}
\end{figure*}

The effectiveness of “CT” can also be seen from the examples. Figures~\ref{result_cityval_using_gta5} and \ref{result_cityval_using_synthia} show that tail-class objects inside dashed boxes are segmented better than DISE after “CT” is performed. In the first row of Figure~\ref{result_cityval_using_gta5}, the traffic sign is segmented better than DISE when “InCT” is applied. Furthermore, when “OutCT” is also applied, the traffic sign’s border and the pole’s border become clearer because “OutCT” enables more reliable supervision.

\section{Conclusion}
In this paper, we proposed the zero-style loss that reduces the domain gap between the source data and target data by completely separating the content and style of the image. The zero-style loss aligns the content features of the two domains, and this leads to improved semantic segmentation performance in the target domain. Moreover, our proposed content transfer enhances performance by solving the class imbalance problem in unsupervised domain adaptation (UDA) for semantic segmentation. In the two UDA segmentation scenarios from synthetic to real, we achieve the state-of-the-art performance.

\section{Acknowledgment}
This research was supported by the National Research Foundation of Korea (NRF) grant funded by the Korea government (MSIT) (NRF-2019R1A2C1007153).

\bigskip
\bibliography{reference.bbl}

\end{document}